\documentclass[lettersize,journal]{IEEEtran}
\usepackage{amsmath,amsfonts}
\usepackage{multirow}
\usepackage{booktabs}
\usepackage{setspace}
\usepackage{algorithm}
\usepackage{algorithmic}
\usepackage{soul}
\usepackage{array}
\usepackage[caption=false,font=normalsize,labelfont=rm,textfont=rm]{subfig}
\usepackage{textcomp}
\usepackage[table]{xcolor}
\usepackage{tabularray}
\usepackage{comment}
\usepackage{stfloats}
\usepackage{url}
\usepackage{verbatim}
\usepackage{colortbl}
\usepackage{graphicx}
\usepackage{cite}

\newsavebox\CBox
\def\textBF#1{\sbox\CBox{#1}\resizebox{\wd\CBox}{\ht\CBox}{\textbf{#1}}}

\def\BibTeX{{\rm B\kern-.05em{\sc i\kern-.025em b}\kern-.08em
    T\kern-.1667em\lower.7ex\hbox{E}\kern-.125emX}}
\usepackage{balance}
\begin{document}
\title{Divide, Ensemble and Conquer: The Last Mile on Unsupervised Domain Adaptation for Semantic Segmentation}
\author{Tao Lian, 
Jose L. G\'{o}mez, 
and Antonio M. L\'opez, \IEEEmembership{Member, IEEE} 
\thanks{Tao Lian, Jose L. G\'{o}mez and Antonio M. L\'opez are with the Computer Vision Center (CVC) at UAB, 08193 Bellaterra (Barcelona), Spain.}
\thanks{
Antonio M. L\'opez are also with the Dpt. of Computer Science, Universitat Aut\`onoma de Barcelona (UAB), 08193 Bellaterra (Barcelona), Spain. 
}
\thanks{The authors acknowledge the support received from the Spanish grant Ref.
PID2020-115734RB-C21 funded by MCIN/AEI/10.13039/501100011033. Antonio M. López
acknowledges the financial support to his general research activities given by ICREA under the
ICREA Academia Program. Tao Lian acknowledges the financial support from the China Scholarship Council (CSC). The authors acknowledge the support of the Generalitat de Catalunya CERCA Program and its ACCIO agency to CVC’s general activities.}
}

\markboth{Journal of \LaTeX\ Class Files,~Vol.~18, No.~9, September~2020}%
{How to Use the IEEEtran \LaTeX \ Templates}

\maketitle

\begin{abstract}
The last mile of unsupervised domain adaptation (UDA) for semantic segmentation is the challenge of solving the syn-to-real domain gap. Recent UDA methods have progressed significantly, yet they often rely on strategies customized for synthetic single-source datasets (e.g., GTA5), which limits their generalisation to multi-source datasets. Conversely, synthetic multi-source datasets hold promise for advancing the last mile of UDA but remain underutilized in current research. Thus, we propose DEC, a flexible UDA framework for multi-source datasets. Following a divide-and-conquer strategy, DEC simplifies the task by categorizing semantic classes, training models for each category, and fusing their outputs by an ensemble model trained exclusively on synthetic datasets to obtain the final segmentation mask. DEC can integrate with existing UDA methods, achieving state-of-the-art performance on Cityscapes, BDD100K, and Mapillary Vistas, significantly narrowing the syn-to-real domain gap.
\end{abstract}
\section{Introduction}
\label{sec:intro}
\soulregister\cite7
\IEEEPARstart{S}{emantic} segmentation \cite{csurka:2022} is a key task in autonomous driving \cite{janai:2020cvadsurvey, tampuu:2020etesurvey} as it provides a detailed understanding of a vehicle's surroundings by assigning a specific class to each pixel in an image. This capability enables the perception module of an autonomous vehicle to identify and distinguish objects such as pedestrians, vehicles, traffic signs, and obstacles to allow safe driving. Training deep learning models for this task requires extensive datasets with precise labels. However, pixel-level human labelling for semantic segmentation is complex and time-consuming to obtain. An approach for addressing this challenge lies in adopting synthetic datasets, wherein automated procedures generate labels, obviating the necessity for manual annotations. Nevertheless, a notable issue arises due to the domain gap – a feature distribution disparity between synthetic datasets' images and real-world scenes. This discrepancy frequently leads to a decline in performance when models trained on synthetic datasets (source domain) are applied to real-world datasets (target domain). A common practice is to employ domain adaptation techniques \cite{wang:2018survey} such as Semi-Supervised Learning (SSL) \cite{yang:2022survey} and Unsupervised Domain Adaptation (UDA) \cite{wilson:2020survey, csurka:2022visual} to bridge the syn-to-real gap. In this paper, we focus on UDA, where the challenge posed by real-world target domains is addressed solely by leveraging synthetic data as the source domain. More specifically, we aim to bridge the last mile, the gap between UDA and supervised learning (SL) with human labelling.
\begin{figure}[t]
    \centering
    \includegraphics[width=1\linewidth]{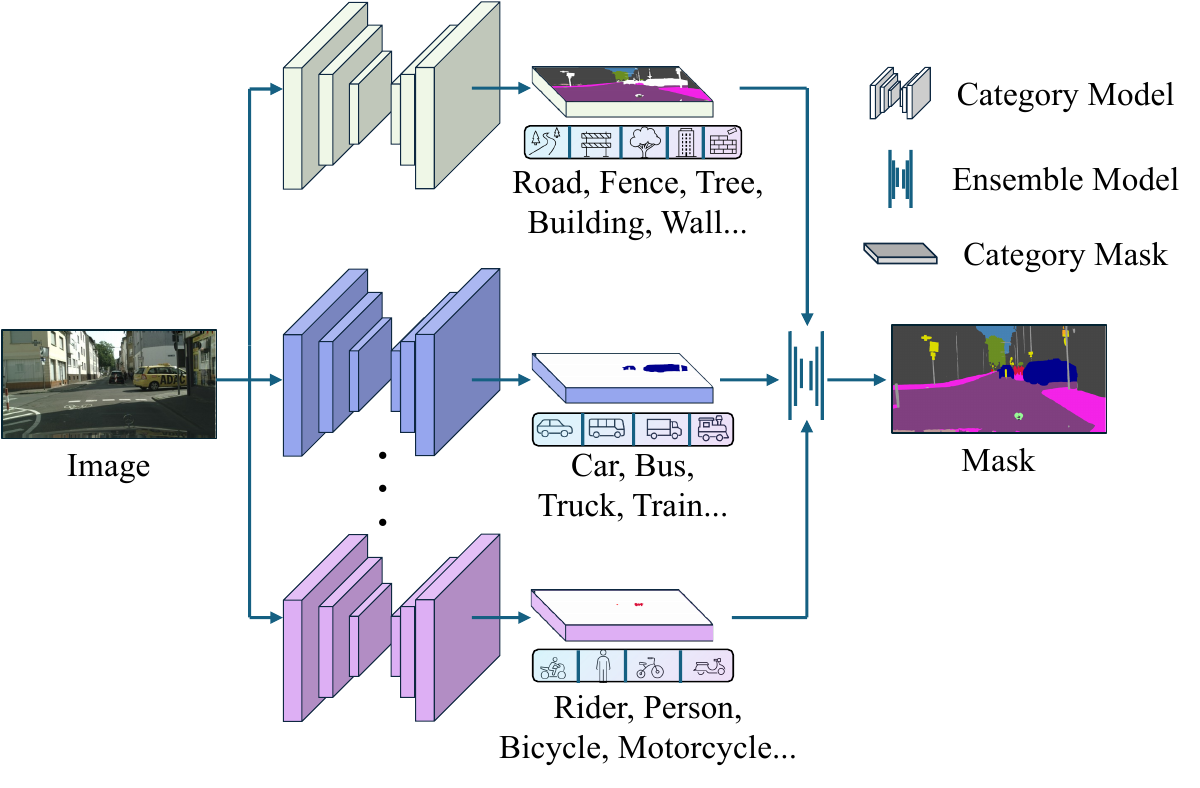}
    \caption{The overview of DEC. It consists of multiple category models and an ensemble model. Category models segment the image into category masks containing distinct classes. Subsequently, the ensemble model fuses these category masks to generate the final mask.}
    \label{fig:DEC}
\end{figure}

The properties of synthetic datasets, such as label accuracy, sample diversity and class balance, significantly impact UDA performance. Most UDA methods \cite{tranheden:2021dacs, hoyer:2022daformer, hoyer:2022hrda, hoyer:2023mic, chen:2023pipa} focus only on popular single-source synthetic domains, such as GTA5 \cite{richter:2016GTA5} or SYNTHIA \cite{ros:2016} datasets; and only targets one single real-world domain, generally Cityscapes \cite{cordts:2016}. Thus, novel strategies such as  Rare Class Sampling (RCS) and Thing-class ImageNet Feature Distance (FD) \cite{hoyer:2022daformer} are designed to address the training challenges associated with these specific source domains that suffer from class imbalance, lack of variability and domain shift. However, these strategies add complexity, instability and overfitting. Alternatively, some works \cite{gomez:2023co-train, gomez:2023urbanSyn, DDB, zhao2019multi} have shown a multi-source approach can successfully replace the necessity of the aforementioned single-domain strategies. In effect, combining several synthetic datasets as one for training solves the class imbalance and increases the variability, improving the generalization capacity of the models. The results from multi-source methods demonstrate a notable reduction in the gap between synthetic and real-world domains. Nevertheless, the multi-source domain has not been adopted extensively since most UDA methods are usually built upon prior research of a single-source domain.

In this paper, we propose a novel multi-source framework called Divide, Ensemble and Conquer (DEC). It follows a divide-and-conquer strategy to train semantic segmentation models and fuse their predictions to generate the semantic segmentation mask. In our first step, we apply the division strategy to train several models on categorized semantic classes, called category models, similarly to how humans semantically label an RGB image using AI tools \cite{Acuna:2018, Bhavani:2023} (first background, then large objects, and finally small ones). The categorization simplifies the number of classes learnt by each category model, expecting a better performance and fast convergence. The next step involves the training of an ensemble model, a DeepLabv3+ \cite{chen:2017DeepLab} network, which learns to fuse segmentation masks generated by each category model into one mask. The framework is illustrated in Fig. \ref{fig:DEC}. DEC exhibits compatibility with a wide range of UDA methods, providing consistent improvement. To ensure a fair comparison, we reproduce the results of previous UDA methods with the multi-source dataset. DEC achieves state-of-the-art performance on Cityscapes, BDD100K \cite{yu:2020BDD}, and Mapillary Vistas \cite{neuhold:2017mapillary}, resulting in mIoU of 78.7, 62.4, and 74.8, respectively. We enhance the existing state-of-the-art reported in \cite{hoyer:2023mic} by 1.2, 2.1, and 0.6 points, respectively. Our performance is only 2.8 and 4.7 points behind SL for Cityscapes and Mapillary Vistas, surpassing the SL for BDD by 0.9 points. Overall, the main contributions of our work are as follows:
\begin{enumerate}
    \item A novel division approach: DEC categorizes semantic classes into groups (e.g., background, large objects, and small objects) and trains category models. This approach simplifies the feature space for each category model, leading to improved UDA performance.
    \item An ensemble-based fusion mechanism: DEC leverages an ensemble model to fuse the outputs from category models into the final segmentation mask. This fusion approach ensures robustness and compatibility with various UDA methods, providing consistent improvements across datasets.
    \item State-of-the-art performance on multiple benchmarks: DEC achieves competitive results on Cityscapes, BDD100K, and Mapillary Vistas, surpassing existing UDA methods and narrowing the gap with SL.
    \item Flexibility and ease of use: Besides the label division strategy, DEC does not rely on specialized strategy or additional training parameters. It can be easily integrated with various UDA methods without changing the baseline model architecture.
\end{enumerate}
\section{Related work}
\label{sec:rel_work}
\subsection{Ensemble Learning}
Ensemble learning is a machine learning approach in which multiple models are trained to solve the same problem and combined to improve performance. One of the first works on CNNs for semantic segmentation using ensembles is Marmanis et al. \cite{marmanis:2016} modifying a FCN network to improve the deconvolution step and then train several networks with different initialisation and average their predictions. Much of the subsequent work focused on changing the structure of the model to apply ensemble learning. Bousselham et al. \cite{bousselham:2021} propose a self-ensemble approach using the multi-scale features produced by a spatial pyramid network to feed different decoders and compose an ensemble by different strategies (averaging, majority vote, and hierarchical attention). Similarly, Cao et al. \cite{cao:2022} use multiple semantic heads sharing the same backbone to compute an ensemble using cooperative learning.  
These approaches further push the performance but increase complexity and lose flexibility. Every model in these works is customised for their algorithms and can not be combined with others easily. Compared with these previous works, \emph{our proposal} is purely data-driven and only requires adapting the labels. This characteristic makes DEC flexible, enabling it to be used with any semantic segmentation model without additional modifications. 

In addition, ensemble learning has been applied to solve UDA tasks. Extensive works are addressed on image classification \cite{rakshit:2019, zhou:2021domain, ahmed:2022, wu:2022}. However, our interest falls in syn-to-real UDA for semantic segmentation where available works are scarcer. Piva et al. \cite{piva:2021} propose a framework similar to \cite{li:2019b} where an image translation module feeds three different semantic segmentation networks, and an ensemble layer aggregates the information to generate pseudo-labels. Chao et al. \cite{chao:2021} propose an end-to-end ensemble-distillation framework that employs different UDA methods and semantic segmentation networks to generate the final pseudo-labels by a pixel and channel-wise fusion policy. The aforementioned works employ various data augmentation techniques, UDA methods, and semantic segmentation networks on each ensemble member. We draw inspiration from these works, emphasising that members' diversity enhances ensemble robustness. \emph{Our proposal} differentiates itself by assigning different classes to ensemble members (divide-and-conquer strategy). Previous studies \cite{chao:2021,bousselham:2021} use majority voting and averaging to ensemble outputs from members, requiring members to be trained with the same classes. To handle category-specific models that output different classes, we propose an ensemble model (CNN) that learns to combine these outputs.

Ensemble learning enhances the generalization performance of deep learning models by combining the predictions of multiple models, reducing variance and mitigating overfitting. This approach enables us to decompose the semantic segmentation task into smaller, more manageable subtasks, each focused on a specific set of semantic categories. By training specialized models for each category, we can leverage the strengths of individual models (category experts), resulting in more robust and accurate segmentation result.
\subsection{UDA for Semantic Segmentation}
Most UDA works relies on single-source synthetic dataset \cite{luo:2019taking, qin:2019generatively, choi:2019self, tsai:2019domain, tuan:2019ADVENT, Lv:2020cross, kim:2020learning, wang:2020differential, wang:2020classes, gao:2021DSP, zheng:2021rectifying, hoyer:2022daformer, hoyer:2022hrda, hoyer:2023mic, li:2019bidirectional, zou:2019confidence, subhani:2020learning, chao:2021}. Hoyer et al. \cite{hoyer:2022daformer} successfully use the novel vision-transformers \cite{dosovitskiy:2021} to solve UDA for semantic segmentation. They employ a teacher-student self-training framework with additional training strategies, such as computing ImageNet feature distances. Furthermore, several new works propose techniques to improve performance on top of existing frameworks (DACS, Daformer, etc.)
Hoyer et al. \cite{hoyer:2022hrda, hoyer:2023mic} adopt a multi-resolution image cropping approach to capture context dependencies and propose a masked image consistency module to learn spatial context relations of the target domain. Chen et al. \cite{chen:2023pipa} explore the pixel-to-pixel and patch-to-patch relation for regularizing the segmentation feature space. These works make significant progress for GTA5 $\rightarrow$ Cityscapes but always need to design specific strategies (e.g., RCS and FD) for class unbalance due to the drawback of GTA5. 
These specific strategies do not generalize effectively to multi-source datasets (see Section \ref{sec:source-data}).
On the other hand, \cite{hoyer:2022daformer,hoyer:2022hrda, hoyer:2023mic, chen:2023pipa} add extra pipelines for training on top of previous works, which makes the UDA become more and more complex and hard to train, e.g., \cite{hoyer:2023mic} is on the top of DACS\cite{tranheden:2021dacs}, Daformer\cite{hoyer:2022daformer} and HRDA\cite{hoyer:2022hrda}. Our DEC method is entirely data-driven and does not require an extra algorithm for UDA training.

Advancements in synthetic image photo-realism have led to the proposal of higher quality datasets than GTA5, e.g., Synscapes \cite{wrenninge:2018synscapes} and UrbanSyn \cite{gomez:2023urbanSyn}. 
UrbanSyn is a novel synthetic dataset that proves effective for UDA. 
Thus, there are a few works on UDA proposing a multi-source approach and tested in different real target datasets \cite{gong:2021mDALU, zhao:2019multisource, he:2021multisource, gomez:2023co-train,gomez:2023urbanSyn}. One of the most promising works with this setup, Gomez \cite{gomez:2023co-train}, proposes an offline co-training method that is purely data-driven and combines two synthetic sources. The multi-source datasets are superior to single-source dataset in terms of scene richness, detailed description, and class balance. These works motivate \emph{our proposal} to focus on multi-source datasets, in contrast to other UDA proposals.
\section{Method}
\begin{figure*}[!t]
    \centering
    \includegraphics[width=1\linewidth]{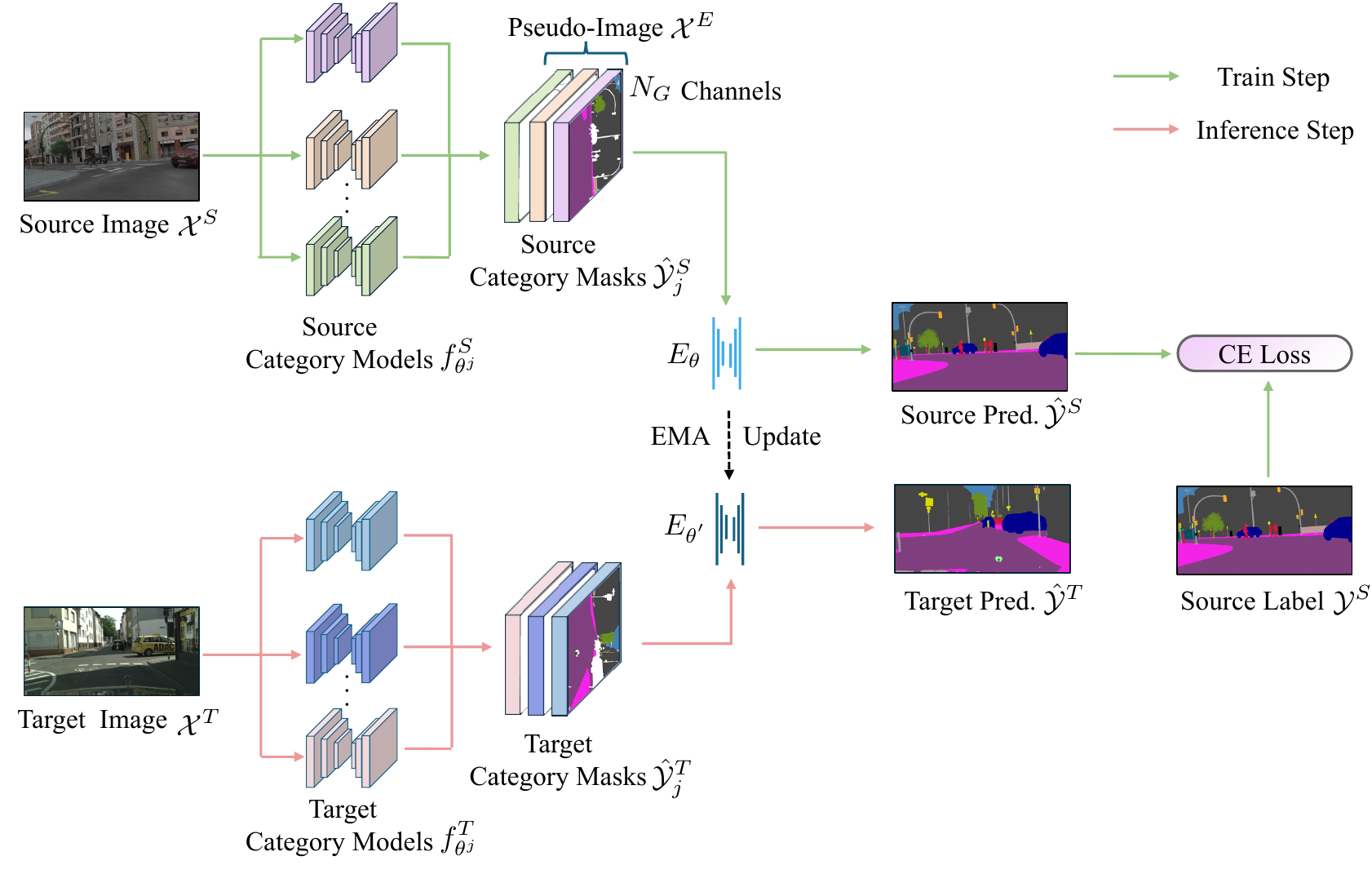}
    \caption{The training procedure for the ensemble model begins by utilising source category models $f_{\theta^j}^S$ to generate source category masks $\hat{\mathcal{Y}}_j^S$. Subsequently, these source category masks $\hat{\mathcal{Y}}_j^S$ are stacked into a pseudo-image $\mathcal{X}^E$ with $N_G$ channels. 
    The student model $E_\theta$ is then trained with source category masks and the source label, and the teacher model $E_{\theta^{'}}$ is updated using the exponential moving average of $E_\theta$ after each training step. For inference, we employ target category models $f_{\theta^j}^T$ to generate target category masks and fuse them into a segmentation mask. It is crucial to note that $f_{\theta^j}^S$ and $f_{\theta^j}^T$ represent distinct models. $f_{\theta^j}^S$ is trained through SL with synthetic dataset, whereas $f_{\theta^j}^T$ is via UDA methods.
}
    \label{fig:train_and_infer}
\end{figure*}
\label{sec:method}
This section details the proposed framework, DEC, grounded in the divide-and-conquer strategy. The framework comprises two components: category models and an ensemble model. Section. \ref{sec:4.1} discusses the division strategy and category models, which segment images into category masks. Subsequently, Section. \ref{sec:4.2} explains the ensemble model, designed to fuse predictions derived from category models.
\subsection{Division Strategy}
\begin{algorithm}[!t]
    \setstretch{1.2}
    \caption{Generate Source Category Label}
    \label{alg:category}
	\renewcommand{\algorithmicrequire}{\textbf{Input:}}
	\renewcommand{\algorithmicensure}{\textbf{Output:}}
	\begin{algorithmic}[1]
	\REQUIRE Set of source label $\mathcal{Y}^S$, a category (list) contains $N_j$ classes, total classes $N_C$
        \ENSURE Source category labels $\mathcal{Y}^S_j$
	\FOR {$i=1$ to $N_C$}
	\IF{$i$ in category}
        \STATE $\mathcal{Y}^S[\mathcal{Y}^S==i] = category.index(i)$
        \ELSE
        \STATE $\mathcal{Y}^S[\mathcal{Y}^S==i] = N_j-1$\\ \COMMENT{The value $N_j-1$ denotes the class \textit{other category}}
        \ENDIF
	\ENDFOR
	\end{algorithmic}
\end{algorithm}
\begin{algorithm}[!t]
    \setstretch{1.2}
    \caption{Train Ensemble Model}
    \label{alg:fusion}
	\renewcommand{\algorithmicrequire}{\textbf{Input:}}
	\renewcommand{\algorithmicensure}{\textbf{Output:}}
	\begin{algorithmic}[1]
	\REQUIRE Set of source image $\mathcal{X}^S$, set of source label $\mathcal{Y}^S$, ensemble model $E_\theta$, source category models $f_{\theta^j}^S$ with $j \in [1..N_G]$
        \ENSURE The trained ensemble model $E_{\theta^{'}}$
	\STATE Set iteration numbers $N$, learning rate $\eta$ and EMA coefficient $\alpha$
	\FOR {$t=1$ to $N$}
	\STATE Choose mini-batch $(x^S, y^S)$
        \STATE Generate source category masks: $\hat{y}^S_j \leftarrow f_{\theta^j}^S(x^S)$
	\STATE Stack category masks: $x^E \leftarrow \sum\limits_{j=1}^{N_G}\hat{y}_j^{S}$
	\STATE Normalize: $x^E=\displaystyle\frac{x^E-\min{(x^E)}}{\max{(x^E)}-\min{(x^E)}} $
    \vspace{0.5em} 
    \STATE Compute prediction: $\hat{y}^E \gets E_\theta(x^E)$
    \vspace{0.5em} 
    \STATE Compute loss: $L \gets \mathcal{H}_{CE}(\hat{y}^E,y^S)$
    \STATE Compute gradients by backpropagation: $\nabla_{\theta} L$
    \STATE Update student model: $\theta_t = \theta_t - \eta \cdot \nabla_{\theta} L$
    \STATE Update teacher model: $\theta_t^{'}=\alpha \theta_{t-1} + (1-\alpha)\theta_t$
	\ENDFOR
	\end{algorithmic}
\end{algorithm}
\label{sec:4.1}
Divide-and-conquer is an algorithmic paradigm where a problem is recursively broken down into smaller related sub-problems that are easier to solve. The solutions to these sub-problems are then combined to address the original problem. In the UDA for semantic segmentation task, a semantic segmentation model $f_\theta$ is trained with source images $\mathcal{X}^S$ and source labels $\mathcal{Y}^S$ to segment target images $\mathcal{X}^T$ into masks $\hat{\mathcal{Y}}^T$ comprising $N_C$ classes. 
\begin{equation}
f_\theta(\mathcal{X}^T)=\hat{\mathcal{Y}}^T, \text{where} \; \hat{\mathcal{Y}}^T\in[0..N_C-1]
\end{equation}
Employing the divide-and-conquer strategy, we deconstruct the semantic segmentation task into subtasks that group classes into categories and train category models to segment different classes. In particular, we group $N_C$ classes into $N_G$ categories guided by specific criteria
which encompasses various factors, such as how humans semantically label an RGB image by AI tools\cite{Acuna:2018, Bhavani:2023} (e.g. commencing with the background, progressing to large objects, and concluding with smaller ones). To account for non-category classes (classes that do not belong to a particular category), we introduce a class named \textit{other category} to denote them. It means a specific class is assigned only in one category label for a given pixel, while other category labels maintain the designation as \textit{other category}. This class helps category models learn features of non-category classes in training, improving the accuracy of classes that belong to a category. Subsequently, we remap source labels $\mathcal{Y^S}$ to source category labels $\mathcal{Y}^S_j$ based on the defined $N_G$ categories where the class will be retained in the category label if it belongs to this category.
The procedure of generating category labels is shown in Algorithm \ref{alg:category}. 
After generating source category labels, we train target category models, denoted as $\{f_{\theta^j}^T\}_{j=1}^{N_G}$, using existing UDA methods.
Each target category model $f_{\theta^j}^T$ is then employed to segment target images $\mathcal{X}^T$ into target category masks $\hat{\mathcal{Y}}^T_j$, encompassing $N_j$ classes.
\subsection{Ensemble Model}
\label{sec:4.2}
By division strategy detailed in Section. \ref{sec:4.1}, target images $\mathcal{X}^T$ are segmented into target category masks $\hat{\mathcal{Y}}^T_j$. Conflicts arise when different classes are predicted for the same pixel value among the target category masks. Note that popular ensemble methods like majority voting and averaging are not suitable to solve these conflicts. Due to the aforementioned division strategy, category models do not share classes, thus ensemble methods that depends on a consensus between models can not be applied. Hence, we introduce an ensemble model $E_\theta$ designed to extract features from each target category mask $\hat{\mathcal{Y}}^T_j$ and fuse them into the final semantic segmentation mask $\hat{\mathcal{Y}}^T$. 
\begin{equation}
    \hat{\mathcal{Y}}^T=E_\theta({\{\hat{\mathcal{Y}}^T_j}\}_{j=1}^{N_G})
\end{equation}
As the ensemble model exclusively fuse masks and yields output identical to the semantic segmentation model, we opt for an encoder-decoder architecture with a simple backbone as the architecture of the ensemble model (see Section. \ref{sec:exp}).

\noindent\textbf{Input}. Given that UDA for semantic segmentation only has labels in the source dataset, for enhanced robustness, we utilise source category masks $\{\hat{\mathcal{Y}}^S_j\}_{j=1}^{N_G}$ and their corresponding source labels $\mathcal{Y}^S$ as the training dataset, denoted as $(\{\hat{\mathcal{Y}}^S_j\}_{j=1}^{N_G},\mathcal{Y}^S)$, to train the ensemble model. Source category masks $\hat{\mathcal{Y}}^S_j$ are generated by source category models $f^S_{\theta^j}$, which are trained on the source dataset.
\begin{equation}
    \hat{\mathcal{Y}}^S_j=f^S_{\theta^j}(\mathcal{X^S}), \text{where} \; j \in [1..N_G]
\end{equation}
Note that source category models $f^S_{\theta^j}$ only need to be trained by SL in contrast to UDA methods since there is no domain gap between source category masks $\hat{\mathcal{Y}}^S_j$ and target category masks $\hat{\mathcal{Y}}^T_j$, both of them are segmentation masks instead of RGB images.

\noindent\textbf{Training Step}. The training step is illustrated in Fig. \ref{fig:train_and_infer} and implemented in Algorithm \ref{alg:fusion}. In the training dataset $(\{\hat{\mathcal{Y}}^S_j\}_{j=1}^{N_G},\mathcal{Y}^S)$, there are $N_G$ category masks for one source label. To enable the model to process this kind of input, we stack them (similar to how we treat RGB images) into a pseudo-image $\mathcal{X}^E$ where each channel corresponds to a source category mask $\hat{\mathcal{Y}}^S_j$. The $\mathcal{X}^E$ channels follow the same order as the $N_G$ categories defined.
We use pixel-wise cross-entropy as the loss function $\mathcal{L}^E$ to train the ensemble model like semantic segmentation task
\begin{equation}
\mathcal{L}^E =\mathcal{H}_{CE}(E_\theta(\mathcal{X}^E),\mathcal{Y}^S)
\end{equation}
and leverage the exponential moving average (EMA) \cite{tarvainen2017ema} to compute the mean of previous model parameters for weight updates, enhancing the robustness and temporal stability of the ensemble model.

\noindent\textbf{Inference Step} The inference step is illustrated in Fig. \ref{fig:train_and_infer}. Like the training step, firstly, we use target category models $f_{\theta_j}^T$ to generate target category masks $\hat{\mathcal{Y}}^T_j$. Then, the trained ensemble model is used to fuse target category masks $\hat{\mathcal{Y}}^T_j$ into the final segmentation mask $\hat{\mathcal{Y}}^T$. 

In this manner, we effectively address the challenge of UDA for semantic segmentation by generating category masks for target images and fusing them into one mask. As target category masks only have edge information, the domain gap between the source and target datasets does not significantly impact the performance of the ensemble model. Consequently, the ensemble model only needs to be trained once and can seamlessly fuse target category masks derived from diverse UDA methods and architectures. This flexibility allows target category models $f_{\theta^j}^T$ to be trained with varied UDA methods and architectural designs, such as Convolutional Neural Networks (CNNs) and Transformers. The efficacy of the ensemble model is demonstrated across various target domains, including Cityscapes, BDD100K, and Mapillary Vistas. Moreover, the output of the ensemble model can serve as pre-annotations for the target dataset, facilitating human annotation or the generation of pseudo-labels for other UDA tasks.

\section{Experiments}
\label{sec:exp}
\subsection{Datasets}
We employ the composite source dataset from \cite{gomez:2023urbanSyn} called Musketeers, which is composed of GTA5 \cite{richter:2016GTA5}, Synscapes \cite{wrenninge:2018synscapes} and UrbanSyn datasets \cite{gomez:2023urbanSyn}. The combination of diverse synthetic datasets results in a more balanced class distribution, ensuring an ample supply of samples, particularly for rare classes. Specifically, GTA5 comprises 24,966 images with a resolution of 1914$\times$1052, Synscapes consists of 25,000 images at 1440$\times$720, and UrbanSyn encompasses 7,539 images with a resolution of 2048$\times$1024. We validate our framework using three distinct datasets as the target domains to show the generalisation capability. Cityscapes includes 2,975 training and 500 validation images, each with a resolution of 2048$\times$1024. BDD100K comprises 7,000 training and 1,000 validation images at resolution 1280$\times$720. The Mapillary Vistas dataset, consisting of 18,000 training and 2,000 validation images, exhibits varying aspect ratios and resolutions. For Mapillary Vistas, we utilise 14,716 training and 1,617 validation images with a 4:3 ratio to ensure consistency. All experiments use the mean intersection over union (mIoU) over all classes as the evaluation metric.
\subsection{Implementation Details}
\subsubsection{Division Strategy}
\begin{figure}[!t]
    \centering
    \includegraphics[width=1\linewidth]{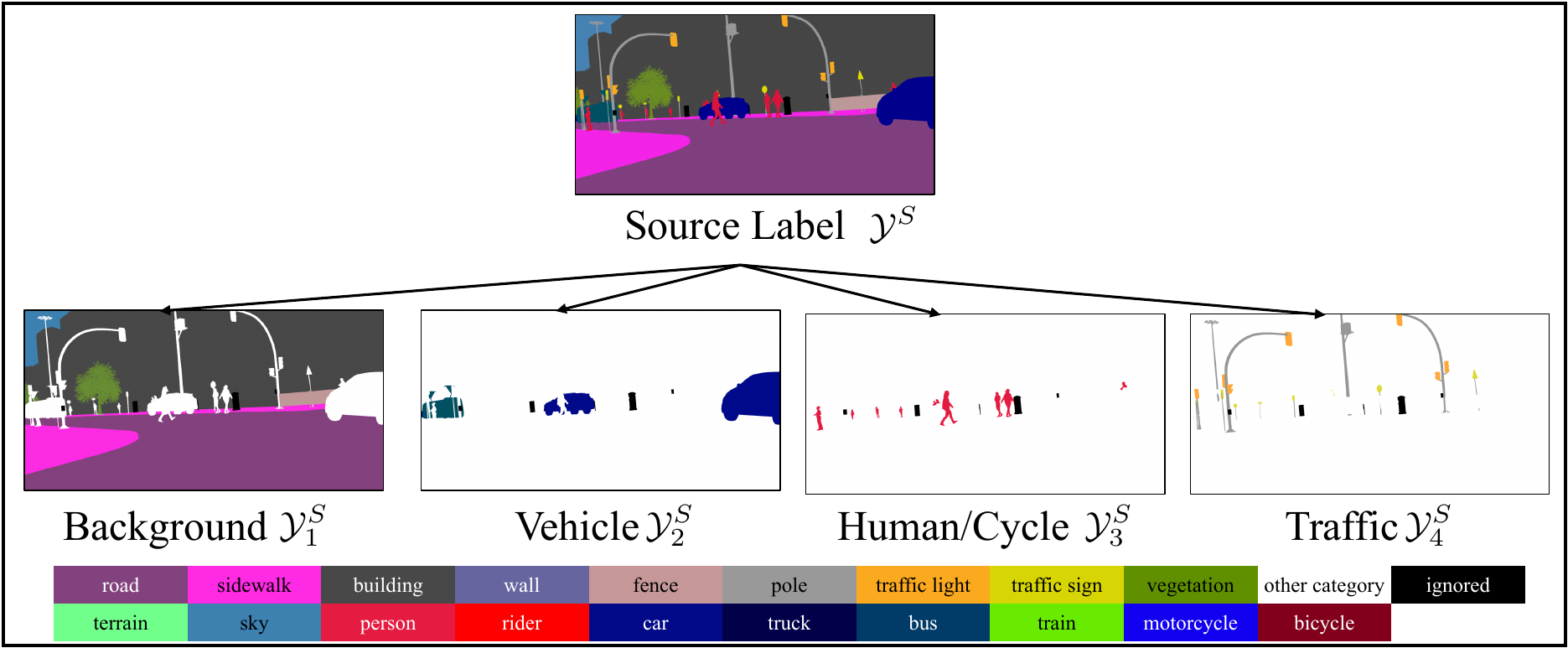}
    \caption{The remapping of a source semantic segmentation label $\mathcal{Y}^S$ to category labels $\{\mathcal{Y}_1^S,\mathcal{Y}_2^S,\mathcal{Y}_3^S,\mathcal{Y}_4^S\}$. For example, in the remapping to Background, only classes belonging to Background are kept; others are marked as \textit{other category} (the white part in the visualisation of Background label).}
    \label{fig:cate-label}
\end{figure}

As a common practice, we focus on nineteen classes from Cityscapes evaluation. The nineteen classes are grouped into four categories based on their object size and relationships, as illustrated in Table \ref{tab:group}. Firstly, we group classes without specific shapes into the Background category. Then, classes with similar shapes and large sizes are grouped into the Vehicles category. For Human/Cycle, we combine \textit{bicycle}, \textit{motorcycle}, \textit{person} and \textit{rider} into the same category due to their relationships that the \textit{person} will be classified as a \textit{rider} if they are on a \textit{bicycle} or \textit{motorcycle}. Finally, The \textit{traffic light}, \textit{traffic sign}, and \textit{pole} are grouped together since traffic items are often attached to poles. Fig. \ref{fig:cate-label} shows the visualisation of generating source category labels.
\begin{table}[!t]
\caption{The component of four categories. The class \textit{other category} in each category represents classes that do not belong to this category.}
\label{tab:group}
\setlength{\tabcolsep}{1.5pt}
\resizebox{1\linewidth}{!}{
\begin{tabular}{l|c}
\toprule
Category    & Class \\
\midrule
Background  & \textit{road}, \textit{sidewalk}, \textit{building}, \textit{wall}, \textit{fence}, \textit{vegetation}, \textit{terrain}, \textit{sky}, \textit{other category} \\
Vehicle     & \textit{car}, \textit{truck}, \textit{bus}, \textit{train}, \textit{other category}\\
Human/Cycle & \textit{person}, \textit{rider}, \textit{motorcycle}, \textit{bicycle}, \textit{other category}\\
Traffic     & \textit{traffic light}, \textit{traffic sign}, \textit{pole}, \textit{other category}\\ 
\bottomrule
\end{tabular}
}
\end{table}
\subsubsection{Source Category Models}Source category models are employed to train the ensemble model. The network architecture uses DeepLabV3+ \cite{deeplabv3plus2018} with ResNet101 \cite{He2015resnet} as the backbone, loaded with ImageNet pretrained weights. Training adopts SGD \cite{ruder2016sgd} with a learning rate of 2$\times10^{-3}$, linear learning rate warmup over 1k iterations, and polynomial decay. The model undergoes training on a batch of eight images with a crop size 1024$\times$512 for 90k iterations.
\subsubsection{Ensemble Model}The ensemble model is a DeepLabV3+ network with ResNet18 as the backbone. The backbone is trained from scratch with random initialisation. The training optimiser is AdamW \cite{loshchilov2018adamw} with a learning rate of $5\times10^{-4}$, incorporating linear learning rate warmup over 8k iterations, polynomial decay and $\alpha=0.9999$. The model is trained on a batch of eight images resized to 2048$\times$1024 for 90k iterations. Note all DeepLabV3+ architectures are from Detectron2 framework \cite{wu2019detectron2}.
\subsubsection{Target Category Models}To advance the last mile of UDA for semantic segmentation, we use the previous state-of-the-art UDA method MIC to train target category models. Consequently, the implementation framework, network architecture, hyper-parameters, optimiser, and training strategy mirror those presented in \cite{hoyer:2023mic}. The resolutions for Cityscapes align with those stipulated in \cite{hoyer:2023mic}; For BDD100K, images are resized to 2560$\times$1440; For Mapillary Vista, we crop each image in the training set by one-third along the height direction and resize it to 2048$\times$1024.
\subsection{Comparison with State-of-the-Art}
\begin{table*}[!t]
\centering
\caption{Comparison with state-of-the-art UDA methods. The results are averaged over three random seeds. Musketeers is composed of GTA5, Synscapes and Urbansyn datasets. SL refers to supervised learning based on human annotations with SegFormer\cite{xie2021segformer}. $\Delta$ denotes the difference between specific UDA method and SL.}
\label{tab:sota}
\setlength{\tabcolsep}{1.5pt}
\resizebox{\linewidth}{!}{%
\begin{tabular}{l|ccccccccccccccccccc|c|c}
\toprule
 Method & Road & S.walk & Build. & Wall & Fence & Pole & Tr.Light & Tr.Sign & Veget. & Terrain & Sky & Person & Rider & Car & Truck & Bus & Train & M.cycle & Bicycle & mIoU&$\Delta$ \\
\toprule
\multicolumn{21}{c}{\textbf{Musketeers $\rightarrow$ Cityscapes}} \\
\toprule
SL\cite{gomez:2023co-train} &98.5 & 87.1 & 93.5 & 65.8 & 65.4 & 68.5 & 74.7 & 81.9 & 93.1 & 64.9 & 95.5 & 84.2 & 66.3 & 95.7 & 84.4 & 91.5 & 84.1 & 72.8 & 79.7 & 81.5&\\
\midrule
DACS\cite{tranheden:2021dacs} & 96.3 & 73.7 & 90.5 & 53.5 & 54.4 & 51.4 & 61.6 & 69.6 & 90.1 & 52.9 & 92.6 & 75.4 & 55.1 & 92.7 & 66.8 & 75.6 & 50.5 & 53.8 & 71.5 & 69.9& -11.6 \\
Daformer\cite{hoyer:2022daformer}  & 95.4 & 71.7 & 90.7 & 54.8 & 56.5 & 56.0 & 61.0 & 68.4 & 91.2 & 55.7 & 93.8 & 75.0 & 53.6 & 92.9 & 71.8 & 77.7 & 66.5 & 57.0 & 72.4 & 71.7& -9.8 \\
Co-Training\cite{gomez:2023co-train} & 96.8 & 76.9 & 91.8 & 56.4 & 58.3 & 62.4 & 67.4 & 75.0 & 91.7 & \textbf{57.2} & 94.6 & 82.1 & 62.0 & 94.8 & 82.7 & 90.0 & 76.4 & 65.2 & 75.0 & 76.7& -4.8 \\
HRDA\cite{hoyer:2022hrda}      & 97.0 & 78.1 & 91.8 & 59.0 & 59.3 & 62.6 & 67.9 & 76.0 & 91.6 & 53.6 & 94.6 & 80.3 & 56.1 & 94.9 & 83.4 & 87.8 & 79.1 & 67.7 & 76.6 & 76.7 & -4.8\\
MIC\cite{hoyer:2023mic}       & 97.3 & 80.4 & 91.8 & 57.9 & 61.1 & \textbf{64.2} & 68.5 & \textbf{76.6} & 91.9 & 52.7 & 94.8 & 81.9 & 59.8 & 94.8 & 83.1 & 90.1 & \textbf{83.0} & 66.8 & 76.8 & 77.5 & -4.0\\
Ours      & \textBF{97.9} & \textBF{82.2} & \textBF{92.3} & \textBF{59.9} & \textBF{61.5} & 63.9 & \textBF{69.2} & 75.0 & \textbf{91.9} & 53.4 & \textBF{95.0} & \textBF{83.8} & \textBF{67.9} & \textBF{95.5} & \textBF{87.8} & \textBF{90.5} & 78.5 & \textBF{70.7} & \textBF{78.4} & \textBF{78.7}& -2.8 \\
\toprule
\multicolumn{21}{c}{\textbf{Musketeers $\rightarrow$ BDD100K}} \\
\toprule
SL & 95 & 66.1 & 87.2 & 34.2 & 52.7 & 51 & 56.8 & 57.4 & 87.4 & 53 & 95.8 & 66.4 & 29.7 & 90.7 & 61.3 & 81.8 & 0 & 48.7 & 52.8 & 61.5&\\
\midrule
DACS\cite{tranheden:2021dacs}      & 90.7 & 41.8 & 73.9 & 14.4 & 34.4 & 44.5 & 38.7 & 40.0 & 80.8 & 40.0 & 87.5 & 58.7 & 23.7 & 87.1 & 34.4 & 54.1 & 0.2 & 46.1 & 57.6 & 49.9& -11.6 \\
Co-Training\cite{gomez:2023co-train} & 92.1 & 44.4 & 81.1 & 24.2 & 44.1 & 43.2 & 46.6 & 40.3 & 72.4 & 36.4 & 87.0 & 64.0 & 57.7 & 87.5 & 46.0 & 76.9 & 0.7 & 51.1 & 55.7 & 55.3& -6.2 \\
Daformer\cite{hoyer:2022daformer}  & 92.2 & 51.8 & 81.4 & 29.4 & 39.6 & 46.9 & 50.2 & 52.3 & 83.6 & 44.2 & 92.8 & 62.3 & 46.0 & 86.8 & 43.5 & 70.3 & 0.0 & \textbf{58.5} & 62.8 & 57.6 & -3.9\\
HRDA\cite{hoyer:2022hrda} & 93.4 & 59.1 & 83.2 & \textbf{34.7} & 43.8 & 48.8 & 52.3 & 52.4 & 83.5 & 45.4 & 93.0 & 67.6 & 48.4 & 87.5 & \textbf{51.5} & \textbf{82.1} & 0.0 & 51.8 & 61.8 & 60.0& -1.5 \\
MIC\cite{hoyer:2023mic} & \textbf{93.6} & 62.5 & 83.1 & 34.2 & 43.0 & 50.7 & 50.5 & 54.4 & \textbf{84.9} & \textbf{48.0} & 93.2 & 65.9 & 47.4 & \textbf{87.6} & 50.8 & 81.5 & 0.0 & 53.3 & 61.6 & 60.3& -1.2 \\
Ours & 93.2 & \textBF{64.5} & \textBF{84.8} & 33.1 & \textBF{44.6} & \textBF{55.4} & \textBF{57.9} & \textBF{56.0} & 84.4 & 47.2 & \textBF{93.4} & \textBF{71.1} & \textBF{55.4} & 86.7 & 51.0 & 81.4 & \textbf{0.0} & 57.8 & \textBF{67.1} & \textBF{62.4}& +0.9 \\
\toprule
\multicolumn{21}{c}{\textbf{Musketeers $\rightarrow$ Mapillary}} \\
\toprule
SL&96.8 & 82.1 & 92.2 & 60.9 & 71 & 67.1 & 73 & 83.3 & 92.5 & 75.6 & 98.7 & 81 & 64.9 & 93.9 & 80 & 87.6 & 68.2 & 67.8 & 73.5 & 79.5&\\
\midrule
DACS\cite{tranheden:2021dacs}      & 60.0 & 41.7 & 83.4 & 31.5 & 46.1 & 46.3 & 62.8 & 74.1 & 80.5 & 45.3 & 59.0 & 71.5 & 47.0 & 88.6 & 51.9 & 50.5 & 16.8 & 54.4 & 60.2 & 56.4& -23.1 \\
Daformer\cite{hoyer:2022daformer}  & 89.7 & 48.2 & 85.4 & 42.5 & 45.0 & 48.9 & 60.0 & 70.6 & 84.8 & 53.3 & 96.6 & 73.9 & 58.5 & 86.7 & 57.8 & 63.6 & 42.2 & 51.9 & 61.7 & 64.3& -15.2 \\
Co-Training\cite{gomez:2023co-train} & 91.7 & 55.2 & 87.2 & 42.3 & 56.1 & 54.1 & 66.6 & 74.7 & \textbf{86.2} & 57.8 & \textbf{97.1} & 79.1 & 59.9 & \textbf{92.6} & 70.1 & 75.4 & 32.0 & 61.5 & 68.5 & 68.8& -10.7 \\
HRDA\cite{hoyer:2022hrda}      & 91.8 & 56.4 & 89.1 & 51.7 & 57.1 & 54.5 & 63.9 & 77.0 & 85.5 & 59.0 & 96.4 & 78.9 & 62.6 & 91.9 & 74.5 & 79.8 & 54.4 & 64.6 & 69.2 & 71.5& -8.0 \\
MIC\cite{hoyer:2023mic}       & \textbf{93.8} & \textbf{66.7} & \textbf{89.9} & \textbf{56.5} & 60.1 & 56.3 & 68.7 & \textbf{78.8} & 85.8 & 58.7 & 96.7 & 80.0 & 63.4 & \textbf{92.3} & \textbf{77.8}& \textbf{84.2} & 62.0 & 66.1 & 72.1 & 74.2&-5.3 \\
Ours   & 93.5 & 64.4 & 89.4 & 54.9 & \textbf{60.1} & \textBF{57.3} & \textBF{68.9} & 75.5 & 85.4 & \textBF{63.7} & 96.4 & \textBF{82.4} & \textBF{69.6} & 92.4 & 76.3 & 84.2 & \textBF{65.2} & \textBF{69.3} & \textBF{72.3} & \textBF{74.8}& -4.7 \\
\toprule
\end{tabular}
}
\end{table*}
\begin{figure}[!t]
    \centering
    \includegraphics[width=1\linewidth]{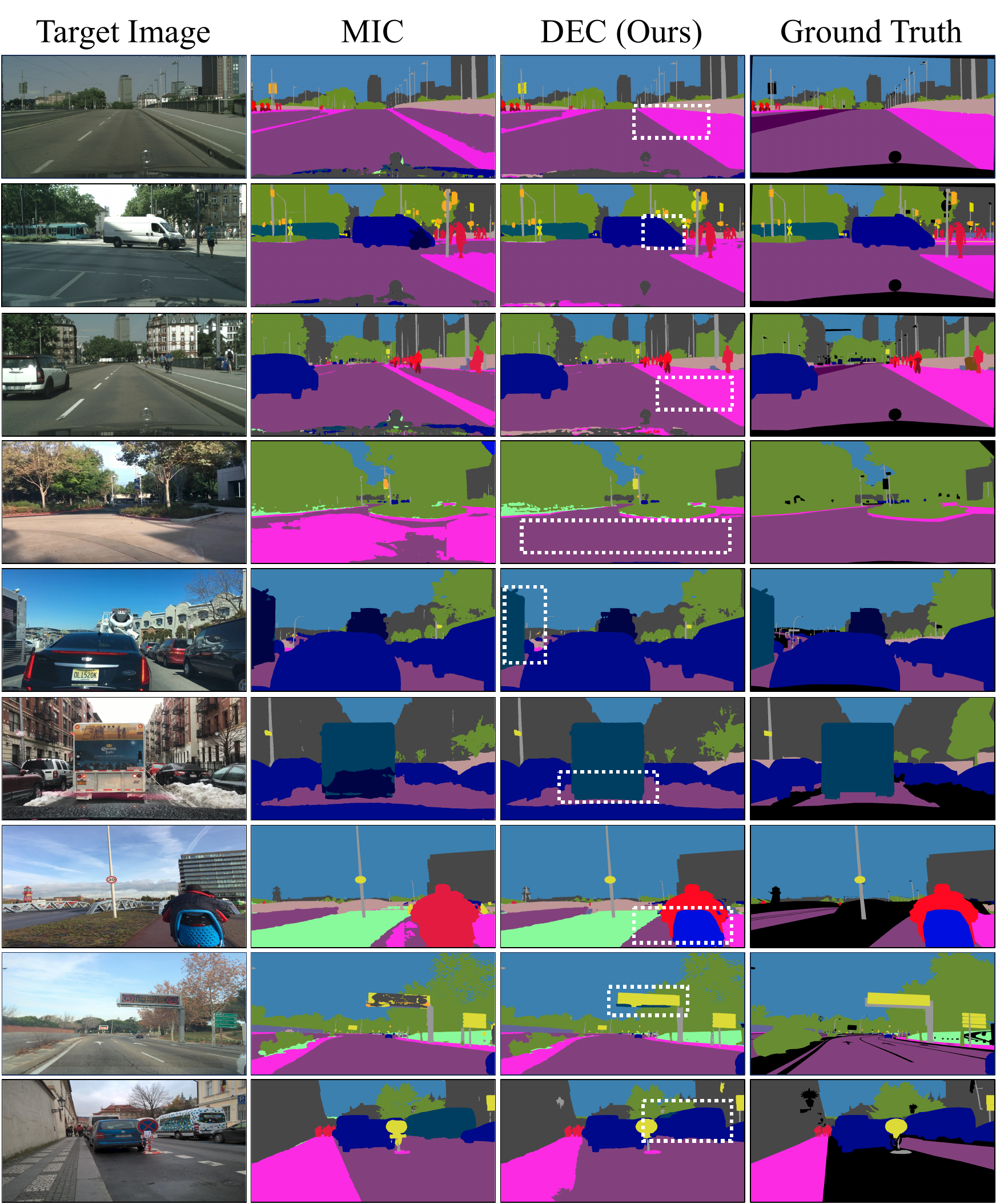}
    \caption{Qualitative results of DEC and previous state-of-the-art method on Musketeers $\rightarrow$ Cityscapes (row 1-3), Musketeers $\rightarrow$ BDD100K (row 4-6) and Musketeers $\rightarrow$ Mapillary Vista (row 7-9). DEC enhances foreground classes, including \textit{traffic sign}, \textit{car}, \textit{motorcycle}, \textit{person}, and \textit{bus}. It effectively achieves better segmentation for \textit{sidewalk}, which is frequently susceptible to confusion with \textit{road}.}
    \label{fig:vis}
\end{figure}
\begin{table*}[!t]
\centering
\caption{DEC with other UDA methods and architectures for Musketeers $\rightarrow$ Cityscapes. The results are averaged over three random seeds. Note the architecture of MIC in this table is DeepLabV2. $\Delta$ denotes the difference with and without DEC.}
\label{tab:other methods}
\setlength{\tabcolsep}{1.5pt}
\resizebox{1\linewidth}{!}{%
\begin{tabular}{lccccccccccccccccccccc}
\toprule
 Method & Road & S.walk & Build. & Wall & Fence & Pole & Tr.Light & Tr.Sign & Veget. & Terrain & Sky & Person & Rider & Car & Truck & Bus & Train & M.cycle & Bicycle & mIoU & $\Delta$ \\
\toprule
DACS\cite{tranheden:2021dacs}       & 96.3 & 73.7 & 90.5 & 53.5 & 54.4 & \textbf{51.4} & 61.6 & \textbf{69.6} & \textbf{90.1} & 52.9 & 92.6 & 75.4 & 55.1 & 92.7 & 66.8 & \textbf{75.6} & \textbf{50.5} & 53.8 & 71.5 & 69.9 & \\
Ours(DACS) & \textbf{97.0} & \textbf{77.2} & \textbf{90.5} & \textbf{57.4} & \textbf{54.7} & 46.6 & \textbf{64.2} & 66.7 & 89.9 & \textbf{55.4} & \textbf{92.7} & \textbf{76.8} & \textbf{58.5} & \textbf{92.9} & \textbf{72.7} & 75.3 & 49.2 & \textbf{58.1} & \textbf{72.7} & \textbf{71.0}& +1.1 \\
\midrule
HRDA\cite{hoyer:2022hrda}       & 97.0 & 78.1 &\textbf{91.8} & 59.0 & \textbf{59.3} & \textbf{62.6} & 67.9 & \textbf{76.0} & 91.6 & 53.6 & 94.6 & 80.3 & 56.1 & 94.9 & 83.4 & 87.8 & \textbf{79.1} & 67.7 & 76.6 & 76.7& \\
Ours(HRDA) & \textbf{97.6} & \textbf{80.2} & 91.6 & \textbf{59.2} & 57.6 & 61.0 & \textbf{68.9} & 74.9 & \textbf{91.7} & \textbf{54.0} & \textbf{94.7} & \textbf{83.8} & \textbf{66.3} & \textbf{95.2} & \textbf{86.4} & \textbf{88.7} & 78.0 & \textbf{68.6} & \textbf{77.6} & \textbf{77.7}&+1.0 \\
\midrule
MIC\cite{hoyer:2023mic} & 96.9 & 76.5 & 90.9 & \textbf{51.3} & 52.4 & 59.3 & 66.7 & 74.1 & 90.9 & 51.7 & 92.9 & 79.7 & 58.3 & 94.0 & 72.4 & \textbf{81.4} & 60.6 & 55.7 & 74.2 & 72.6 &\\
Ours(MIC)       & \textbf{97.1} & \textbf{77.6} & \textbf{91.4} & 51.0 & \textbf{54.2} & \textbf{61.0} & \textbf{69.2} & \textbf{75.2} & \textbf{91.0} & \textbf{52.1} & \textbf{93.3} & \textbf{82.4} & \textbf{64.6} & \textbf{94.2} & \textbf{76.5} & 79.7 & \textbf{62.2} & \textbf{63.9} & \textbf{76.5} & \textbf{74.4}& +1.8\\
\toprule
\end{tabular}
}
\end{table*}
Table \ref{tab:sota} presents the results of our method alongside those of previous state-of-the-art UDA methods. Our framework achieves state-of-the-art performance in Cityscapes, BDD100K and Mapillary Vista. On Cityscapes, we elevate the mIoU from 77.5 to 78.7, a difference of 2.8 points compared to SL. Compared to MIC, which demonstrates a margin of 0.8 points on top of HRDA, our method outperforms MIC with a greater margin of 1.2. Notably, on BDD100K, our method surpasses MIC by 2.1 points and exceeds the SL by 0.9 points. The improvement over SL can be attributed to the challenging nature of this target dataset, which lacks proper balance. Therefore, incorporating a class-rich and balanced synthetic dataset, such as Musketeers, contributes to the superior performance of our method compared to SL. In addition to improving mIoU in nineteen classes, our method particularly enhances performance in foreground classes. For Cityscapes, our approach achieves notable improvements in IoU for \textit{person} with +1.9, \textit{rider} with +8.1, \textit{truck} with +4.7, \textit{motorcycle} with +3.9, and \textit{bicycle} with +1.6. Remarkably, the IoU for \textit{rider} and \textit{truck} exceeds SL by margins of +1.3 and +3.0, respectively. For BDD100K and Mapillary Vista, \textit{person}, \textit{rider}, \textit{motorcycle}, and \textit{bicycle} are also significantly improved, validating the efficacy of our division strategy in enhancing model performance for these foreground classes. Besides quantitative results, we present visualisations comparing our method with the state-of-the-art approach for Musketeers $\rightarrow$ Cityscapes, BDD100K and Mapillary Vistas to illustrate the advancements achieved by our proposed method (see Fig. \ref{fig:vis}).
\subsection{DEC with Other Methods and Architectures}
Our framework can integrate with diverse UDA methods without introducing additional training parameters or strategies. We extend the implementation of our method to DACS with DeepLabV2, MIC with DeepLabV2, and HRDA with Daformer for Musketeers $\rightarrow$ Cityscapes. The corresponding results are detailed in Table \ref{tab:other methods}. Notably, DEC significantly enhances performance across numerous foreground classes, including \textit{rider}, \textit{motorcycle}, and \textit{bicycle}. In addition to foreground classes, DEC demonstrates improvements in background classes such as \textit{sidewalk}, which is often confounded with \textit{road}. Specifically, \textit{sidewalk} improves from 73.7 to 77.2 for DACS, 78.1 to 80.2 for HRDA, and 76.5 to 77.6 for MIC.
\subsection{Ablation Study}
\subsubsection{Source data}
\begin{figure*}[!t]
    \centering
    \includegraphics[width=1\linewidth]{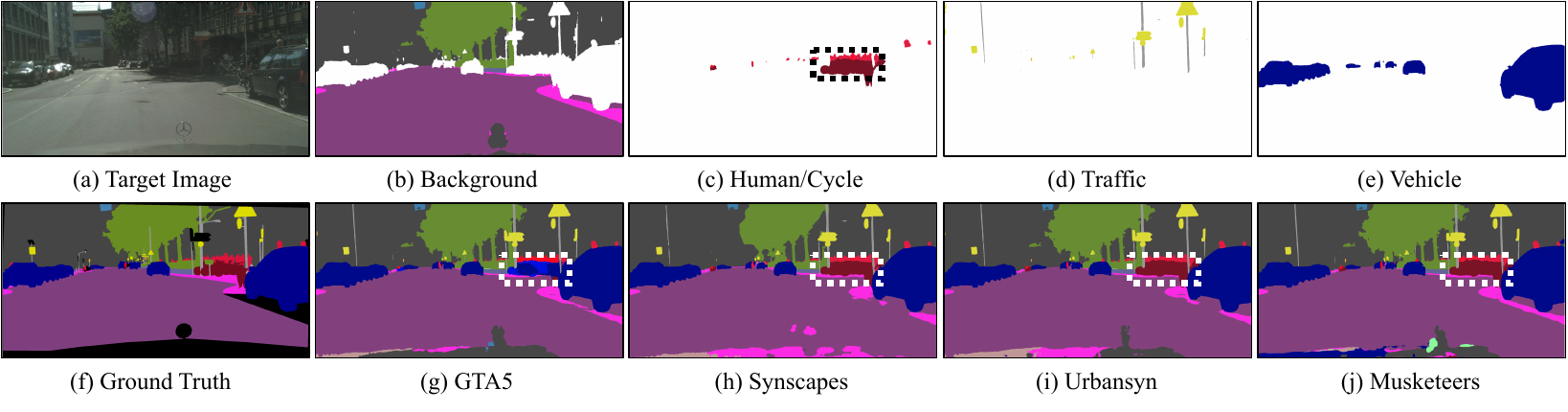}
    \caption{Qualitative results of target category models and different ensemble models on Musketeers $\rightarrow$ Cityscapes. Subfigures (b-e) display predictions from target category models trained on Musketeers, while subfigures (g-j) show predictions from ensemble models trained on GTA5, Synscapes, UrbanSyn and Musketeers.}
    \label{fig:ensembl-vis}
\end{figure*}
\label{sec:source-data}
\begin{table}[!t]
\setlength{\tabcolsep}{5pt}
\centering
\caption{Rare Class Sampling (RCS) and Thing-class ImageNet Feature Distance (FD) are applied to a multi-source dataset, with the results averaged over three random seeds.}
\label{tab:tricks}
\resizebox{1\linewidth}{!}{
\begin{tabular}{cccccc}
\toprule
Method                &Source& Target            & w/o RCS, FD & w RCS, FD & Improvement \\
\midrule
\multirow{3}{*}{MIC}  &\multirow{3}{*}{Musketeers}& Cityscapes   & 77.5      & 76.8      & -0.7        \\
                      & & BDD100K           & 60.3        & 59.7      & -0.6        \\
                      & & Mapillary   Vista & 74.2        & 73.6      & -0.6        \\
\midrule                  
\multirow{3}{*}{HRDA} &\multirow{3}{*}{Musketeers}& Cityscapes   & 76.7        & 75.7      & -1.0        \\
                      & & BDD100K           & 60.1        & 58.8      & -1.3        \\
                      & & Mapillary   Vista & 71.7        & 71.6      & -0.1        \\
\bottomrule
\end{tabular}
}
\end{table}
\begin{table}[!t]
\centering
\setlength{\tabcolsep}{25pt}
\caption{The influence of single-source and multi-source training datasets on the ensemble model. We show the performance of GTA5 $\rightarrow$ Cityscapes and Musketeers $\rightarrow$ Cityscapes on the top of MIC.}
\label{tab:ensemble single}
\resizebox{1\linewidth}{!}{
\begin{tabular}{ccc}
\toprule
Target Category Models & Ensemble Model & mIoU \\
\midrule
\multirow{2}{*}{GTA5}    & GTA5              & 72.6 \\
                      & Musketeers              & \textbf{77.0}   \\
                      \midrule
\multirow{4}{*}{Musketeers}    & GTA5              & 74.6 \\
                      & Synsapes              & 77.3 \\
                      & UrbanSyn              & 78.4 \\
                      & Musketeers            & \textbf{79.2} \\
                      \bottomrule
\end{tabular}
}
\end{table}
DEC use multi-source datasets to train category and ensemble models. For category models, as stated in Section \ref{sec:intro}, strategies tailored for single-source synthetic dataset are often susceptible to overfitting and instability. MIC and HRDA exhibit noteworthy advancements in GTA $\rightarrow$ Cityscapes based on RCS and FD while these strategies demonstrate suboptimal performance in multi-source scenarios (see Table \ref{tab:tricks}). On the other hand, the ensemble model needs enough training data since there is no appropriate pre-trained backbone. The variability of single-source dataset is insufficient for developing a robust ensemble model. Table \ref{tab:ensemble single} presents the mIoU of ensemble models trained on single-source and multi-source datasets. The results demonstrate that the ensemble model trained on multi-source datasets effectively fuses category masks into the final mask, e.g., the Human/Cycle category (see Fig. \ref{fig:ensembl-vis} (c)) correctly segments the dotted area, whereas the ensemble model trained with GTA5 fuse dotted area incorrectly (see Fig. \ref{fig:ensembl-vis} (g)). Thus, it is necessary to train the ensemble model with multi-source datasets to obtain robust and reliable final masks.
\subsubsection{Division Strategy}
\begin{figure}[!t]
    \centering
    \includegraphics[width=1\linewidth]{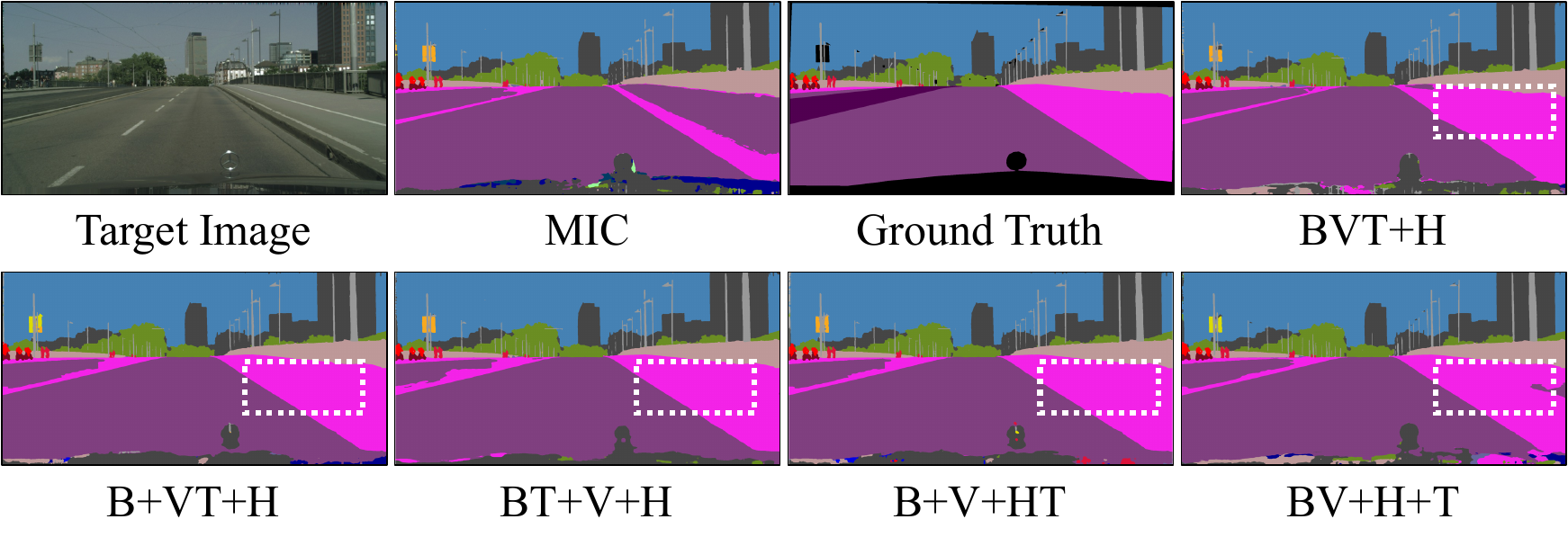}
    \caption{Qualitative results of different feasible division strategies and MIC on Musketeers $\rightarrow$ Cityscapes.}
    \label{fig:different-strategy}
\end{figure}

\begin{table}[!t]
\centering
\caption{Random division strategy. Each row in the second column is a category.}
\label{tab:random division}
\setlength{\tabcolsep}{-5pt}
\resizebox{1\linewidth}{!}{
\begin{tabular}{cc}
\toprule
Random Division                 & Classes                                                                      \\
\midrule
\multirow{4}{*}{1} & Pole, Train, Terrain, Traffic Light, Truck, Bicycle, Road, Wall, other category              \\
                         & Motorcycle, Sky, Person, Building, other category                                            \\
                         & Vegetation, Traffic Sign, Sidewalk, Fence, other category                                    \\
                         & Bus, Rider, Car, other category                                                              \\
                         \midrule
\multirow{4}{*}{2} & Bicycle, Traffic Sign, Person, Fence, Truck, Sidewalk, Car, Traffic Light, other category    \\
                         & Rider, Pole, Building, Terrain, other category                                               \\
                         & Bus, Motorcycle, Wall, Train, other category                                                  \\
                         & Road, Vegetation, Sky, other category                                                        \\
                         \midrule
\multirow{4}{*}{3} & Motorcycle, Pole, Bus, Traffic Sign, Sky, Terrain, Sidewalk, Fence, other category           \\
                         & Car, Train, Truck, Wall, other category                                                      \\
                         & Person, Traffic Light, Rider, Building, other category                                       \\
                         & Vegetation, Bicycle, Road, other category                                                    \\
                         \midrule
\multirow{4}{*}{4} & Rider, Traffic Light, Motorcycle, Car, Truck, Traffic Sign, Pole, Vegetation, other category \\
                         & Wall, Road, Bicycle, Sky, other category                                                     \\
                         & Bus, Person, Train, Building, other category                                                 \\
                         & Sidewalk, Fence, Terrain, other category     
                            \\
                            \bottomrule
\end{tabular}
}
\end{table}
DEC groups classes into categories to decrease the complexity of the model and improve segmentation. However, DEC requires a feasible division strategy (including but not limited to Table \ref{tab:group}) to maintain the contextual information between different classes. Choosing proper categories is not an arbitrary task. We demonstrate in Table \ref{tab:random division} how randomly pick some classes as a group to generate some categories. It is seen that random division does not push the performance for Musketeers $\rightarrow$ Cityscapes in Table \ref{tab:different strategy result} (strategy 1-4). Using the four-category division (strategy B+V+H+T), DEC achieves a +1.2 mIoU improvement over MIC. Additionally, we show the performance of other feasible division strategies, such as dividing classes into two categories and three categories in Table \ref{tab:different strategy result}.
\begin{table*}[!t]
\arrayrulecolor{black}
\caption{Performance of selected, random and other feasible division strategies on Musketeers $\rightarrow$ Cityscapes.}
\label{tab:different strategy result}
\setlength{\tabcolsep}{1.5pt}
\resizebox{1\linewidth}{!}{
\begin{tabular}{c|cccccccccccccccccccc}
\toprule
Strategy & Road & S.walk & Build. & Wall & Fence & Pole & Tr.Light & Tr.Sign & Veget. & Terrain & Sky & Person & Rider & Car & Truck & Bus & Train & M.cycle & Bicycle & mIoU \\
\midrule
\multicolumn{21}{c}{Without Division}\\
\midrule
MIC & 97.3 & 80.4 & 91.8 & 57.9 & 61.1 & 64.2 & 68.5 & 76.6 & 91.9 & 52.7 & 94.8 & 81.9 & 59.8 & 94.8 & 83.1 & 90.1 & 83.0 & 66.8 & 76.8 & 77.5 \\
\midrule
\multicolumn{21}{c}{With Random Division}\\
\midrule
1  & 96.7 & 76.5   & 92.6  & 60.5 & 59.6  & \textbf{65.8 }& \textbf{70.3 }         & 76.9         & 92.2  & 56.4    & 95.3 & 82.1   & 56.8  & 95.1 & 84.4  & 84.8 & 70.7  & 67.7    & 72.8    & 76.7 \\
2  & 90.6 & 55.7   & \textbf{92.7}  & 60.0 & 61.1  & 64.8 & 68.5          & 75.8         & 92.1  & 52.0    & \textbf{95.4} & 80.9   & 51.3  & 95.1 & 87.3  & 88.7 & 81.9  & 63.7    & 76.1    & 75.5 \\
3  & 91.3 & 58.0   & 92.5  & 58.6 & 59.6  & 65.3 & 69.1          & 77.0         & 91.9  & 54.7    & 95.0 & 83.8   & 66.3  & 95.0 & 84.9  & 87.7 & 77.9  & 65.0    & 73.7    & 76.2 \\
4  & 90.3 & 55.6   & 92.4  & 57.1 & 56.9  & 65.5 & 70.1          & 76.3         & \textbf{92.3}  & 53.8    & 95.1 & 82.2   & 62.3  & 95.1 & 86.5  & 88.1 & 80.3  & 69.1    & 63.6    & 75.4 \\
\midrule
\multicolumn{21}{c}{With B+V+H+T Division}\\
\midrule
B+V+H+T      & \textBF{97.9} & 82.2 & 92.3 & 59.9 & 61.5 & 63.9 & 69.2 & 75.0 & 91.9 & 53.4 & 95.0 & 83.8 & 67.9 & 95.5 & 87.8 & 90.5 & 78.5 & \textbf{70.7} & 78.4 & 78.7 \\
\midrule
\multicolumn{21}{c}{With Other Feasible Division}\\
\midrule
BV+HT    & 97.4 & 79.9   & 92.1  & 61.4 & 61.1  & 64.6 & 68.9          & \textbf{77.1}         & 92.1  & 54.4    & 95.1 & 83.3   & 63.1  & 94.9 & 80.5  & \textbf{91.6} & 82.4  & 69.2    & 77.9    & 78.3 \\
BT+VH    & 97.6 & 81.1   & 92.4  & 61.8 & 60.6  & 64.5 & 68.2          & 75.7         & 92.1  & 56.1    & 94.9 & 83.1   & 63.2  & 95.4 & 86.2  & 90.9 & 78.8  & 67.1    & 77.8    & 78.3 \\
B+V+HT   & \textbf{97.9} & 82.8   & 92.6  & 59.9 & \textbf{62.3}  & 65.4 & 69.0          & 76.3         & 91.9  & 54.2    & 95.0 & 82.8   & 62.8  & 95.4 & 87.4  & 91.4 & 79.8  & 67.6    & 77.3    & 78.5 \\
B+VT+H   & \textbf{97.9} & \textbf{82.9}   & 92.5  & 59.8 & \textbf{62.3}  & 64.5 & 68.0          & 76.1         & 91.9  & 54.3    & 95.0 & \textbf{84.4}   & \textbf{68.0}  & 95.3 & 86.1  & 89.6 & 76.6  & 70.5    & 78.3    & 78.6 \\
BV+H+T   & 97.8 & 81.8   & 92.0  & 60.4 & 60.4  & 64.1 & 68.9          & 74.9         & 92.1  & 54.5    & 95.0 & 84.3   & 67.9  & 94.9 & 80.2  & 91.5 & 83.4  & 70.4    & \textbf{78.5}    & 78.6 \\
BT+V+H   & 97.7 & 81.1   & 92.4  & 62.0 & 60.7  & 64.4 & 68.0          & 75.4         & 92.0  & 56.3    & 94.9 & \textbf{84.4}   & 67.7  & \textbf{95.6} & \textbf{87.9}  & \textbf{91.6} & 80.4  & 70.1    & 78.2    & 79.0 \\
BVT+H    & 97.8 & 82.2   & 92.2  & \textbf{64.3} & 60.5  & 64.2 & 68.9          & 73.2         & 92.1  & \textbf{56.5}    & 95.2 & \textbf{84.4}   & 67.7  & 95.3 & 86.8  & 91.2 & \textbf{83.8}  & \textbf{70.7}    & \textbf{78.5}    & \textbf{79.2} \\
\bottomrule
\end{tabular}
}
\end{table*}
These division strategies are generated by combining categories in Table \ref{tab:group}, e.g., \textit{BV} denotes a category containing classes from Background and Vehicle; \textit{BV+HT} means this division strategy splits all classes into two groups. These division strategies achieve a gain of at least $+0.8$ for Musketeers $\rightarrow$ Cityscapes compared to MIC. Fig. \ref{fig:different-strategy} provides a qualitative result where these feasible strategies can effectively improve segmentation, e.g., \textit{sidewalk}.
\subsubsection{Ensemble Model}
We choose the neural network DeepLabV3+ as an encoder-decoder architecture to ensemble the output of our category models. This network is well-known, easy to train and does not require a lot of resources. Here, we demonstrate the influence of different backbones and training parameters (e.g. learning rate and EMA) on the ensemble model.

\begin{table}[!t]
\caption{The performance of ensemble model with different backbones for three target domains.}
\centering
\label{tab:backbone}
\setlength{\tabcolsep}{18pt}
\resizebox{1\linewidth}{!}{
\begin{tabular}{lccc}
\toprule
Backbone  & Cityscapes & BDD100K & Mapillary Vista \\
\midrule
ResNet18  & \textbf{79.1}       & \textbf{63.2} & \textbf{75.0} \\
ResNet34  & 78.9       & 63.1 & 74.7 \\
ResNet50  & 78.2       & 62.6 & 74.2 \\
ResNet101 & 78.9       & 63.1 & \textbf{75.0} \\
\bottomrule
\end{tabular}
}
\end{table}
\noindent\underline{Backbone:} Table \ref{tab:backbone} illustrates the performance of the ensemble model with various backbones for Musketeers → Cityscapes, BDD100K and Mapillary Vista. ResNet-18 achieves the highest mIoU across all three datasets (79.1, 63.2, and 75.0, respectively). Given its superior performance and fastest training speed among the tested backbones, ResNet-18 is selected as the preferred backbone for the ensemble model.

\begin{table}[!t]
\centering
\caption{The performance of SGD and AdamW optimiser with different learning rates for Musketeers $\rightarrow$ Cityscapes.}
\setlength{\tabcolsep}{1.5pt}
\label{tab:optimizer}
\resizebox{1\linewidth}{!}{
\begin{tabular}{ccccc|cccc}
\toprule
Optimizer & \multicolumn{4}{c}{SGD}           & \multicolumn{4}{c}{AdamW}         \\
\midrule
Learning Rate        & $5\times10^{-2}$ & $5\times10^{-3}$ & $5\times10^{-4}$ & $5\times10^{-5}$ & $5\times10^{-2}$ & $5\times10^{-3}$ & $5\times10^{-4}$ & $5\times10^{-5}$ \\
\midrule
mIoU      & 74.7   & 78.2   & 76.6   & 61.4   & 78.9   & \textbf{79.1}   & \textbf{79.1}   & 78.6  \\
\bottomrule
\end{tabular}
}
\end{table}
\noindent\underline{Optimizer:} Tab. \ref{tab:optimizer} shows the SGD and AdamW with different learning rates to train the ensemble model for Musketeers $\rightarrow$ Cityscapes. It can be seen that AdamW is more compatible with the ensemble model than SGD.

\begin{table}[!t]
\caption{The impact of EMA coefficient $\alpha$ for three target domains.}
\centering
\label{tab:ema}
\setlength{\tabcolsep}{18pt}
\resizebox{1\linewidth}{!}{
\begin{tabular}{rccc}
\toprule
$\alpha$ &Cityscapes & BDD100K & Mapillary Vista \\
\midrule
0.9                     & \textbf{79.1} & 62.9 & 74.8 \\
0.99                    & 75.1 & 59.8 & 71.6 \\
0.999                   & 78.7 & 62.8 & 74.6 \\
0.9999                  & \textbf{79.1} & \textbf{63.2} & \textbf{75.0}  \\
\bottomrule
\end{tabular}
}
\end{table}
\noindent\underline{EMA:} Tab. \ref{tab:ema} shows the influence of different EMA coefficients $\alpha$ for Musketeers $\rightarrow$ Cityscapes, BDD100K and Mapillary Vista. It is observed that a larger value of $\alpha$ enhances the generalisation of the ensemble model, leading to optimal performance for each target dataset.
\subsection{Performance Metrics}
\begin{table}[!t]
\setlength{\tabcolsep}{15pt}
\centering
\caption{FPS, FLOPs and Parameter size for Input Resolution 1024$\times$512 on RTX 3090}
\label{table:fps_flops_params}
\resizebox{\linewidth}{!}{%
\begin{tabular}{lccc}
\toprule
                 & FPS   & FLOPs & Params \\
                 &    &  (GFLOPs) &  (M) \\ \midrule
HRDA                   & 4.50  & 410.40         & 85.69      \\
Category Model (HRDA)  & 4.50  & 410.40         & 85.69      \\
\midrule
MIC                    & 4.50  & 410.40         & 85.69      \\
Category Model (MIC)   & 4.50  & 410.40         & 85.69      \\
\midrule
Ensemble Model         & 122.24 & 37.40          & 12.30      \\ \bottomrule
\end{tabular}
}
\end{table}
\begin{table}[!t]
\setlength{\tabcolsep}{4pt}
\centering
\caption{Runtime and memory consumption during training and inference on RTX 3090}
\label{table:runtime_memory}
\resizebox{\linewidth}{!}{ %
\begin{tabular}{lcccc}
\toprule
                 & \multicolumn{2}{c}{Training} & \multicolumn{2}{c}{Inference (2048$\times$1024)} \\ \cmidrule(lr){2-3} \cmidrule(lr){4-5}
                 & Throughput & GPU Memory & Throughput & GPU Memory \\
                 & (s/iter) & (GB) &  (img/s) &  (GB) \\\midrule
HRDA                   & 2.55            & 21.80           & 1.13              & 11.93          \\
Category Model (HRDA)  & 2.35            & 15.07           & 1.13              & 11.93          \\ \midrule
MIC                    & 3.13            & 21.73           & 1.13              & 11.93          \\
Category Model (MIC)   & 2.70            & 15.09           & 1.13              & 11.93          \\
\midrule
Ensemble Model         & 0.69            & 17.89           & 46.30              & 0.98           \\ \bottomrule
\end{tabular}
}
\end{table}
Table \ref{table:fps_flops_params} presents the FPS, FLOPs, and parameters for both the category and ensemble models. The category model is built on the existing UDA method, with modifications made to the training labels. Consequently, its FPS, FLOPs, and parameters remain the same as those of the HRDA and MIC models. The ensemble model, with only 12.30M parameters, achieves notably high inference speeds, enabling efficient fusion of the category model outputs. As a result, the overall running speed is primarily dictated by the UDA method used to build the category models—the faster the UDA method, the better our approach performs.

The runtime and memory consumption are presented in Table \ref{table:runtime_memory}. The category model achieves faster training and reduced memory usage than HRDA and MIC by eliminating unnecessary techniques, such as rare class sampling and the thing-class ImageNet feature distance. Meanwhile, the inference speed and memory consumption remain the same with HRDA and MIC, as no additional parameters are introduced. The ensemble model's simple architecture completes each training iteration in just 0.69 seconds. Additionally, the ensemble model achieves an inference speed (FPS) of 46.3 for an input resolution of 1024$\times$2048 while consuming only 0.98 GB of memory.
\subsection{The Last Mile to SL}
\begin{figure}[!t]
    \centering
    \includegraphics[width=1\linewidth]{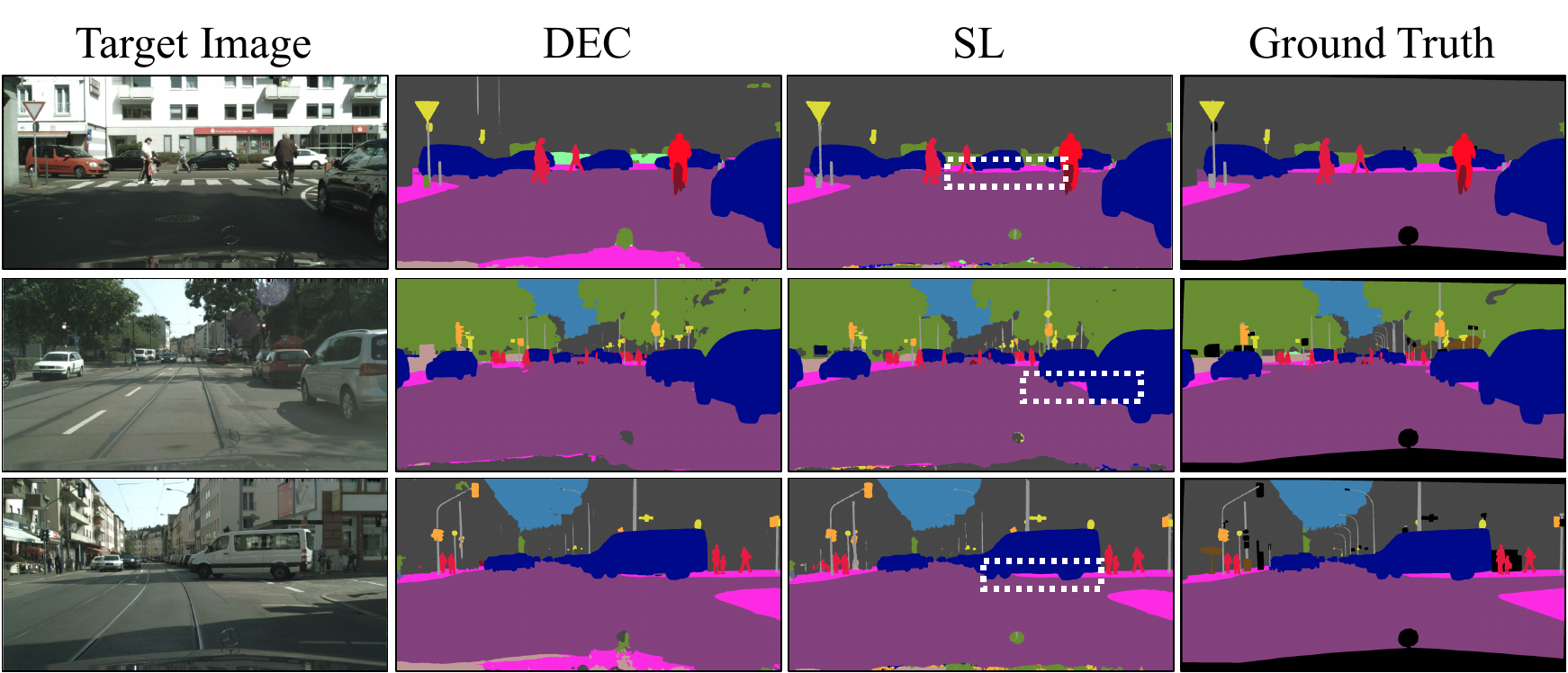}
    \caption{Visualisation of \textit{sidewalk} in DEC and SL on Musketeers $\rightarrow$ Cityscapes.}
    \label{fig:last-mile}
\end{figure}
UDA for semantic segmentation is becoming trustworthy and closing the gap with respect SL. However,  can it reach equal performance? This remaining gap is what we call the last mile of UDA. Our work advances the last mile, with specific target domains (e.g., BDD100K) and some classes (e.g., \textit{rider}, \textit{truck}) even surpassing SL.

For the commonly used target domain, Cityscapes, our DEC method outperforms SL for \textit{rider} and \textit{truck}. Additionally, for the classes \textit{road}, \textit{sky}, \textit{person}, \textit{car}, and \textit{bus}, DEC achieves a difference of less than one point compared to SL. For BDD100K, DEC crosses the last mile, improving performance beyond SL. As for Mapillary, the performance for the classes \textit{person}, \textit{rider}, and \textit{motorcycle} surpasses that of SL, classes that are crucial for autonomous driving.

\begin{figure*}[!t]
    \centering
\includegraphics[width=1\linewidth]{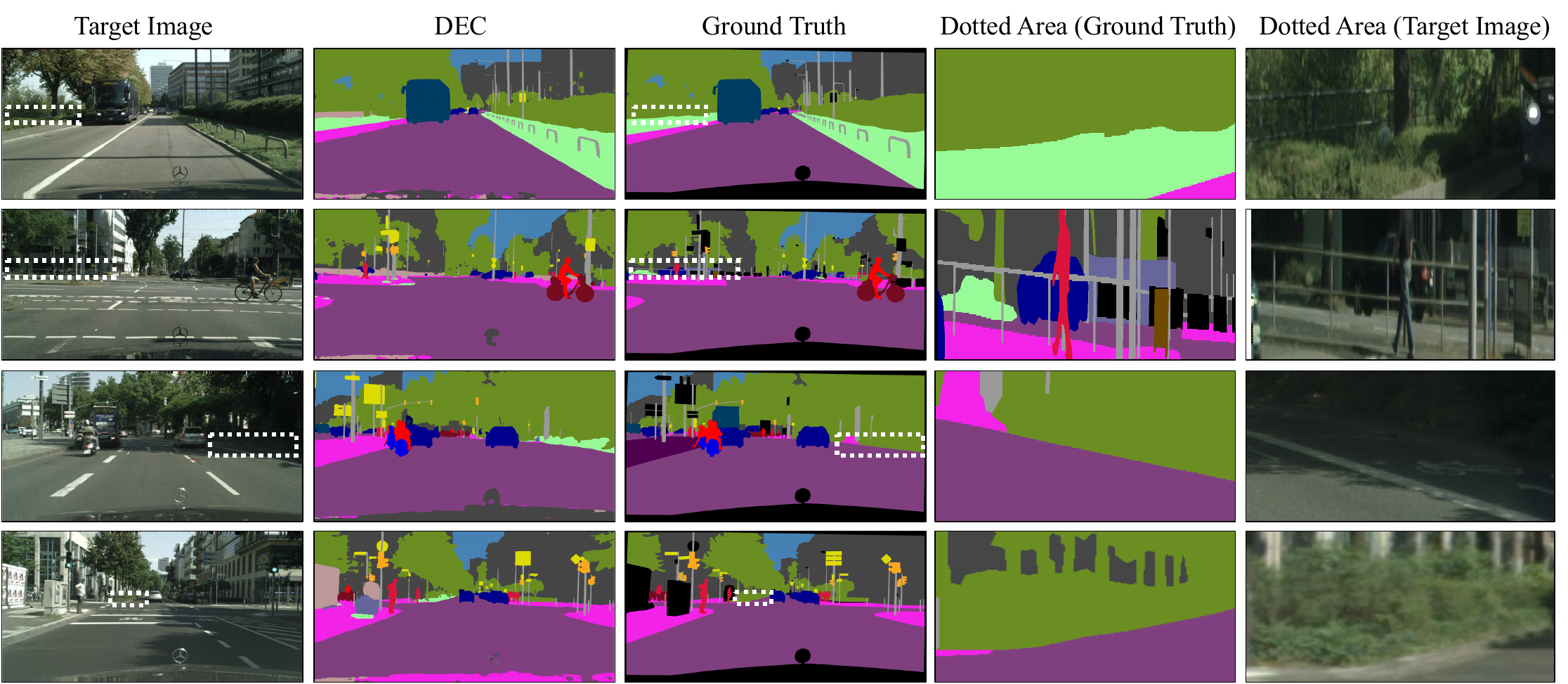}
    \caption{Samples of human bias in the annotation of Cityscapes. The dotted area is labelled in rows 1 and 2 as different classes, \textit{vegetation} and \textit{pole}, respectively. To our eyes, these areas resemble \textit{fence}, as predicted by our DEC method. In rows 3 and 4, according to the definition of classes in Cityscapes, it is very difficult to decide whether the dotted area is \textit{vegetation} or \textit{terrain} because it spreads horizontally and vertically. To our eyes, this area is more likely to be \textit{terrain}, although it is labelled as \textit{vegetation}}
    \label{fig:last-mile-diff}
\end{figure*}

\begin{figure}[!t]
    \centering
\includegraphics[width=1\linewidth]{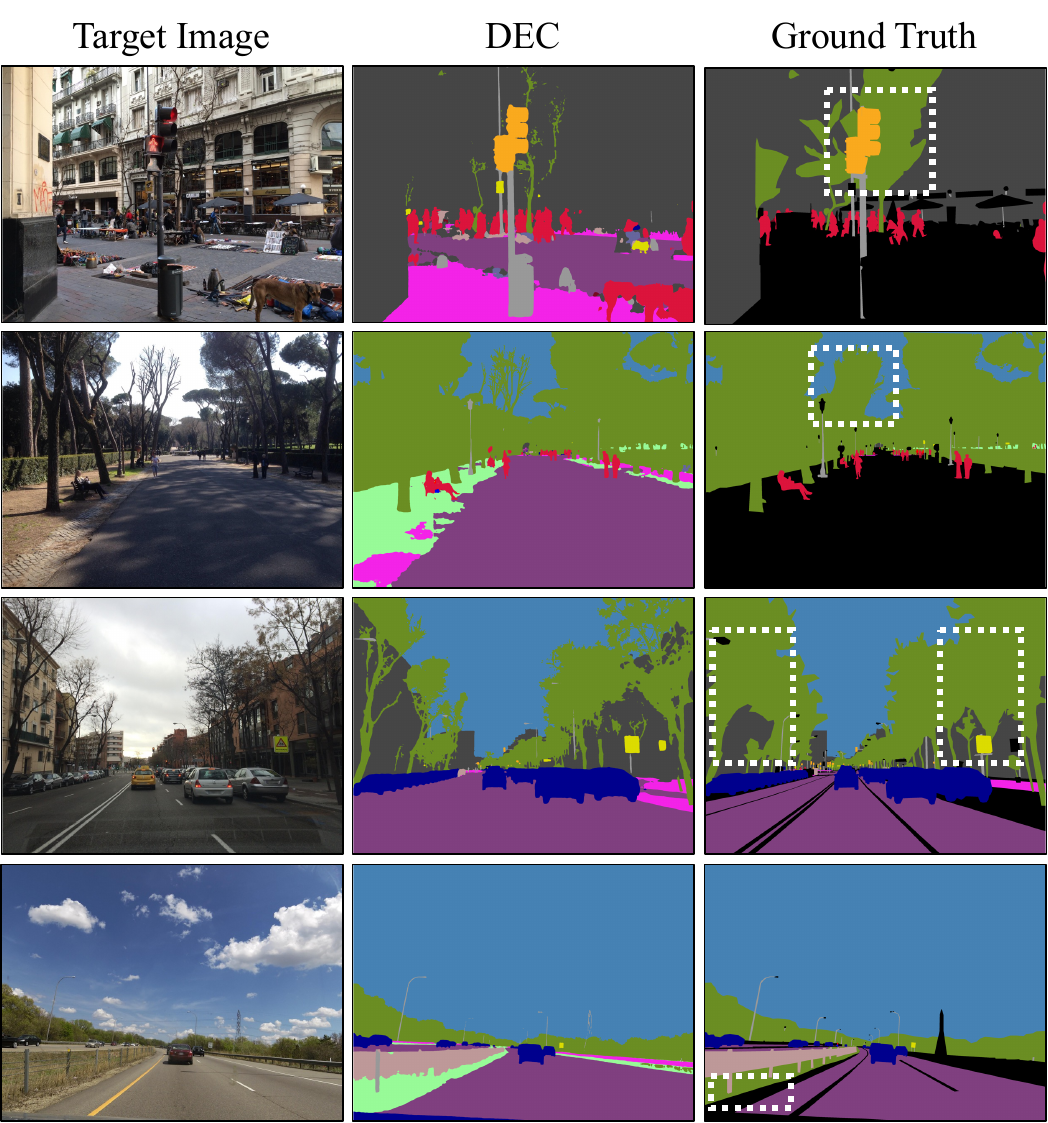}
    \caption{Samples of human bias in the annotation of Mapillary Vista. In rows 1-3, these dotted areas are not finely labelled, with some parts of \textit{building} and \textit{sky} labelled as \textit{terrain}. In row 4, the dotted area is expected to be \textit{terrain} but is labelled as \textit{vegetation}.}
    \label{fig:mapillary-bias}
\end{figure}

Some classes still exhibit gaps compared to SL. For example, the \textit{sidewalk}, an important class on autonomous driving, shows a 5 points difference from SL on Musketeers $\rightarrow$ Cityscapes. Fig. \ref{fig:last-mile} shows some qualitative results for the \textit{sidewalk} in both DEC and SL. The error in DEC is primarily found at the bottom of vehicles, where the \textit{sidewalk} is misclassified as \textit{road}. We expect these errors to have an insignificant influence on autonomous driving, as the vehicles in these areas are classified correctly. Moreover, since the ground truth is manually labelled, human bias is inevitably introduced into the ground truth. This is why some classes show discrepancies compared to SL, such as \textit{terrain}, \textit{fence}, and \textit{pole}. Fig. \ref{fig:last-mile-diff} illustrates these human biases in Cityscapes.
Furthermore, there are also human bias in Mapillary Vistas. Fig. \ref{fig:mapillary-bias} indicates annotation bias among \textit{terrain}, \textit{building}, \textit{sky}, and \textit{vegetation}.
These classes are labelled differently between synthetic and real datasets. This explains the large gap (e.g., \textit{vegetation}, \textit{terrain}) in our method for Mapillary Vista.
\section{Conclusion}
\label{sec:con}
This paper introduces DEC, a pioneering framework utilizing multi-source datasets in category models to address the last mile of UDA for semantic segmentation. DEC comprises category models and an ensemble model, which are responsible for segmenting images into category masks and fusing them into a segmentation mask. Compatible with previous UDA methods, DEC consistently demonstrates improvement when applied on top of them. Across three real datasets, namely Cityscapes, BDD100K, and Mapillary Vistas, DEC achieves state-of-the-art performance with mIoU of 78.7, 62.4, and 74.8, respectively. Compared to SL, DEC lags by merely 2.8 and 4.7 points for Cityscapes and Mapillary Vistas, while it surpasses by 0.9 points for BDD100K.
\bibliographystyle{ieeetr}
\bibliography{main.bbl}
\vfill
\end{document}